\documentclass[letterpaper, 10 pt, conference]{ieeeconf}  % Comment this line out if you need a4paper

\IEEEoverridecommandlockouts                              % This command is only needed if
\usepackage{graphics} % for pdf, bitmapped graphics files
\usepackage{amsmath} % assumes amsmath package installed
\usepackage{amssymb}  % assumes amsmath package installed
\usepackage[colorlinks, linkcolor=blue, citecolor=blue]{hyperref}
\usepackage{cite}
\usepackage{amssymb}
\usepackage{graphicx}
\usepackage{caption}
\usepackage{subfigure}
\usepackage{mathrsfs}
\usepackage{array}
\usepackage{multirow}
\usepackage{makecell}
\usepackage{xcolor}
\usepackage{amsfonts}
\usepackage{gensymb}
\makeatletter
\renewcommand*\env@matrix[1][*\c@MaxMatrixCols c]{%
 \hskip -\arraycolsep
 \let\@ifnextchar\new@ifnextchar
 \array{#1}}
\makeatother

\title{\LARGE \bf
2-Entity RANSAC for robust visual localization in changing environment
}
\author{Yanmei Jiao, Yue Wang, Bo Fu, Xiaqing Ding, Qimeng Tan, Lei Chen and Rong Xiong% <-this % stops a space
\thanks{Yanmei Jiao, Yue Wang, Bo Fu, Xiaqing Ding and Rong Xiong are with the State Key Laboratory of Industrial Control and Technology, Zhejiang University, Hangzhou, P.R. China. Qimeng Tan and Lei Chen are with the Beijing Key Laboratory of Intelligent Space Robotic System Technology and Applications, Beijing Institute of Spacecraft System Engineering, Beijing, P.R. China. Yue Wang is the corresponding author {\tt\small wangyue@iipc.zju.edu.cn}. Rong Xiong is the co-corresponding author.}%
}

\begin{document}

\maketitle
\thispagestyle{empty}
\pagestyle{empty}

%$y_{\mathcal{C}}$
%$p_{c_{1}_{x}}-p_{c_{2}_{x}}$
%$^{\mathcal{C}}p_{c_{3}}}$

%%%%%%%%%%%%%%%%%%%%%%%%%%%%%%%%%%%%%%%%%%%%%%%%%%%%%%%%%%%%%%%%%%%%%%%%%%%%%%%%

\begin{abstract}
Visual localization has attracted considerable attention due to its low-cost and stable sensor, which is desired in many applications, such as autonomous driving, inspection robots and unmanned aerial vehicles. However, current visual localization methods still struggle with environmental changes across weathers and seasons, as there is significant appearance variation between the map and the query image. The crucial challenge in this situation is that the percentage of outliers, i.e. incorrect feature matches, is high. In this paper, we derive minimal closed form solutions for 3D-2D localization with the aid of inertial measurements, using only 2 point matches or 1 point match and 1 line match. These solutions are further utilized in the proposed 2-entity RANSAC, which is more robust to outliers as both line and point features can be used simultaneously and the number of matches required for pose calculation is reduced. Furthermore, we introduce three feature sampling strategies with different advantages, enabling an automatic selection mechanism. With the mechanism, our 2-entity RANSAC can be adaptive to the environments with different distribution of feature types in different segments. Finally, we evaluate the method on both synthetic and real-world datasets, validating its performance and effectiveness in inter-session scenarios.
\end{abstract}

%%%%%%%%%%%%%%%%%%%%%%%%%%%%%%%%%%%%%%%%%%%%%%%%%%%%%%%%%%%%%%%%%%%%%%%%%%%%%%%%
\section{Introduction}

Localization is essential for autonomous robot navigation. Compared to the LiDAR-based localization methods, visual localization is favorable in many applications as the low-cost and stable cameras are widely available. A typical workflow of the visual localization task is to detect the feature points in the query image (FAST \cite{rosten2006machine}, SIFT \cite{lowe2004distinctive}), match the image feature points to the map points (ORB \cite{rublee2011orb}, SIFT \cite{lowe2004distinctive}), and estimate the pose based on a set of matches. Match outliers are hardly avoidable in this setting, leading to inaccurate pose estimation. Random sample consensus (RANSAC) \cite{fischler1981random} is a popular method to achieve robust estimation by randomly sampling the matching set and voting for the inliers. However, RANSAC is limited by serious appearance changes in the environment, in which the percentage of outliers may grow significantly \cite{ding2018laser}\cite{tang2018topological}. Therefore, reliable visual localization robust to the weather, illumination or seasonal changes remains a challenging problem.

\begin{figure}[tp]
      \centering
      \includegraphics[width=0.47\textwidth]{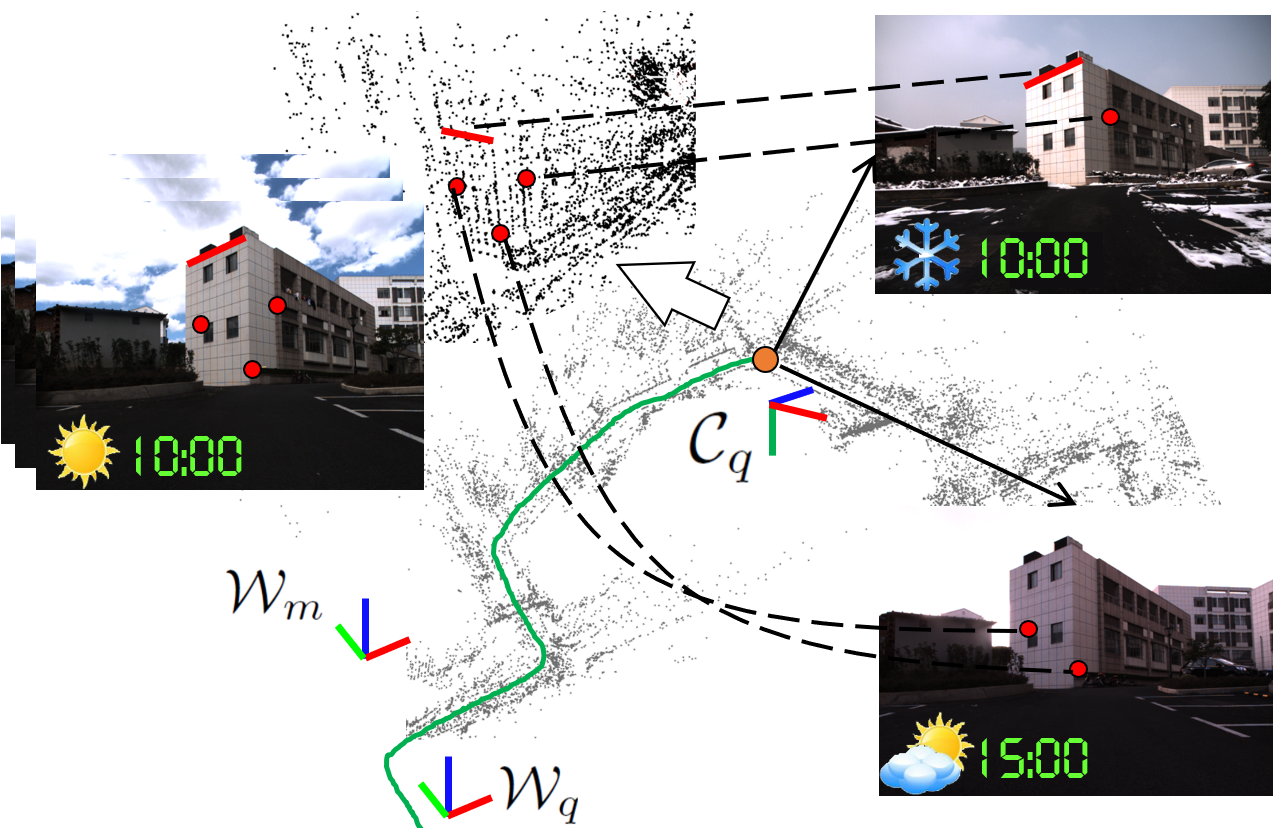}
      \caption{The green line indicates the current session localizing in the black pre-built 3D map with 1 point 1 line or 2 points minimal solution. The map was built in summer, while the query sessions are under different weather or seasons.}
      \label{fig.overview}
   \end{figure}

Two factors are crucial for RANSAC to find inliers, i.e. the number of inliers in the set (higher is better), and the minimal number of matches required to estimate a pose (lower is better). For the first factor, some previous works propose to increase the number of inliers by utilizing multiple types of features (e.g. points, lines and planes)\cite{dornaika1999pose}\cite{raposo2013plane}, as the feature defined on larger image regions usually leads to higher robustness against the illuminative variation. In this paper, we also consider multiple types of features following the practice of previous methods \cite{ramalingam2011pose}. For the second factor, the well-known 3D-2D localization methods typically require 3 or 4 feature matches such as perspective-3-points (P3P) \cite{gao2003complete}, efficient perspective-n-points (EPnP) \cite{lepetit2009epnp}, the minimal solutions of 3 lines \cite{dhome1989determination}\cite{chen1990pose} or a mixture of 3 entities (i.e. points and lines) \cite{ramalingam2011pose}. Our key observation of this paper is that the number of matches can be further reduced for both point and line features in robotics navigation scenario, leading to higher robustness against outliers.

Specifically, with the growing trend of visual inertial navigation in robots, the direction of gravity between the pre-built map and the current query image can be easily aligned. Therefore, we aim to reduce the number of point feature matches from 3 to 2 for closed form pose estimation, by making use of the directly observed pitch and roll angle. Furthermore, the closed form solution is also derived for the case of 1 point and 1 line, so that line feature matches can be considered. As a result, we propose a 2-entity RANSAC for persistent visual localization as illustrated in {Fig. \ref{fig.overview}}. The 2-entity is named for that only two feature matches between points and lines are used to pose estimation. In summary, we present the contributions of the paper as follows:
%  In theory, as 2 observations (2D) can provide constraints for 4 degrees of freedom (DoF), the proposed solution should be minimal.
\begin{itemize}
\item We propose a general framework to derive the closed form solutions to pose estimation which can deal with both point and line features, namely 2 point matches or 1 point match and 1 line match.
\item We propose a 2-entity RANSAC by embedding the minimal closed form solutions into the RANSAC framework and proposed three sampling strategies in RANSAC that select different types of feature matches for pose estimation.
\item We analyze the success probability for different sampling strategies in the 2-entity RANSAC, and propose a selection mechanism that adaptively select strategy depending on environmental characteristics, i.e. structured or unstructured.
\item The proposed method is evaluated on both synthetic and multiple real world data sessions. The results validate the effectiveness and efficiency of our method on visual localization with seasonal and illuminative changes.
\end{itemize}

The remainder of this paper are organized as follows: In Section II we review some relevant visual pose estimation methods. Section III gives the overview of the proposed method and the derivation of the closed form solutions. The analysis about the criterion to select the sampling strategy in RANSAC is addressed in Section IV. We then evaluate our method in Section V on synthetic data and challenging real world data. Finally, the conclusion is drawn in Section VI, which completes this paper.

\section{Related Works}

The minimal solution with efficient 2D-2D point matches for camera pose estimation has been studied extensively for decades. There are many well-known methods, such as 8-point method for fundamental matrix estimation which yields a unique solution \cite{gruen2013calibration}. In addition, there are also 7-point \cite{hartley2003multiple}, 6-point \cite{philip1996non} and 5-point \cite{nister2004efficient} methods which focus on further reduction of minimal number of point matches required for pose estimation. In recent years, with the development of visual inertial navigation, this number is further reduced by considering the measurement of angular velocity and the gravity. Based on this, a 3-point method was proposed in \cite{fraundorfer2010minimal}. More recently, the 2-point RANSAC \cite{troiani20142}\cite{kneip2011robust} was proposed by employing the gyro to calculate the relative rotation within short duration, which was employed in several visual inertial odometry softwares. By further constraining the motion pattern of the camera, only 1 point match is sufficient to derive the pose \cite{scaramuzza2009real}.

In localization, 3D-2D matches are available but there is no measurement of relative yaw angle. In this case, the minimal number of required feature matches is reduced compared with scenario of 2D-2D. Specifically, there are several 3-point methods \cite{gao2003complete}\cite{haralick1994review}\cite{kneip2011novel}\cite{wang2018efficient}\cite{ke2017efficient} that give up to four solutions in closed-form and other 3-line solvers \cite{dhome1989determination}\cite{chen1990pose} that give up to eight solutions which require slower iterative methods. As significant change can appear in long term, it's difficult to find enough point matches or line matches alone to guarantee the success rate of localization. Thus the methods utilizing both point and line features are studied in \cite{ramalingam2011pose} including two possible cases: 2 points 1 line case and 1 point 2 lines case.
With the aid of inertial measurement,the 2-point RANSAC solution is investigated in \cite{kneip2011robust} and \cite{kukelova2010closed}. While the former utilized the relative measurement of yaw angle, which is only available between two consecutive camera under odometry problem, thus cannot be applied in localization, and the latter only considers point feature matches. To the best of our knowledge, the minimal solution for localization using both point and line features with the aid of inertial measurements has not been studied yet, which is the focus of this paper.

\section{Minimal Solutions}
% denote the coordinates $\scriptstyle{_{\mathcal{C}_{q}}^{\mathcal{W}_{m}}}$$T$
The definition of coordinates in this paper is shown in {Fig. \ref{fig.overview}}. A 3D map is built by running a visual inertial simultaneous localization and mapping (VI-SLAM), of which the world reference frame is denoted as ${\mathcal{W}_{m}}$. In the current query session, a visual inertial navigation system (VINS) is utilized with its origin defined as ${\mathcal{W}_{q}}$. Denoting the coordinates of the current query camera as ${\mathcal{C}_{q}}$, the localization problem is stated as the estimation of the query image pose with respect to the world, i.e. $T_{{\mathcal{W}_{m}}{\mathcal{C}_{q}}}$. Thanks to the observability of pitch and roll angle, $\tilde{\beta}$ and $\tilde{\gamma}$, given by the inertial measurements, the direction of gravity can be aligned between ${\mathcal{W}_{m}}$ and ${\mathcal{W}_{q}}$ \cite{li2013high}, leading to the reduction of degrees of freedom (DoF) in pose as
\begin{equation}
  {T_{{\mathcal{W}_{m}}{\mathcal{C}_{q}}}}=[R_{z}(\alpha)R_{y}(\tilde{\beta})R_{x}(\tilde{\gamma})|(\begin{matrix}T_{1} & T_{2} & T_{3}\end{matrix})^{T}] \nonumber
\end{equation}
where $\alpha$ and $(T_{1},T_{2},T_{3})$ denote the yaw angle and three translations to be estimated.  In sequel, for the purpose of clearance, we are going to estimate the inversion of the pose, i.e. ${T_{{\mathcal{C}_{q}}{\mathcal{W}_{m}}}}$. Note that the number of unknown DoFs in ${T_{{\mathcal{C}_{q}}{\mathcal{W}_{m}}}}$ remains the same.

%We denote $R_{mn}$ as the entry in the $m$th row and $n$ column in rotational part and $t_{m}$, the $m$th entry in the translational part of ${T_{{\mathcal{W}_{m}}{\mathcal{C}_{q}}}}$.
%$$
%{T_{{\mathcal{W}_{m}}{\mathcal{C}_{q}}}} \in {\mathbb{R}^{4\times4}} =
%\begin{bmatrix}
%    R & t \\
%    {\bf{0}}_{1\times3} & 1
%  \end{bmatrix}.
%$$

%
%$$ {R_{{\mathcal{W}}{\mathcal{C}}}} \triangleq (R_{z}(\alpha)R_{y}(\tilde{\beta})R_{x}(\tilde{\gamma}))^{T} $$
%$$ {t_{{\mathcal{W}}{\mathcal{C}}}} \triangleq - {(R_{z}(\alpha)R_{y}(\tilde{\beta})R_{x}(\tilde{\gamma}))^{T}} \cdot {(\begin{matrix}t_{1} & t_{2} & t_{3}\end{matrix})^{T}}. $$

%
%Additionally, the m-th row and n-th column entry of the rotation matrix is denoted as $R_{mn}$, the m-th entry of the translation matrix is denoted as $T_{m}$.

\begin{figure}[tbp]
    \centering
    \includegraphics[width=0.48\textwidth]{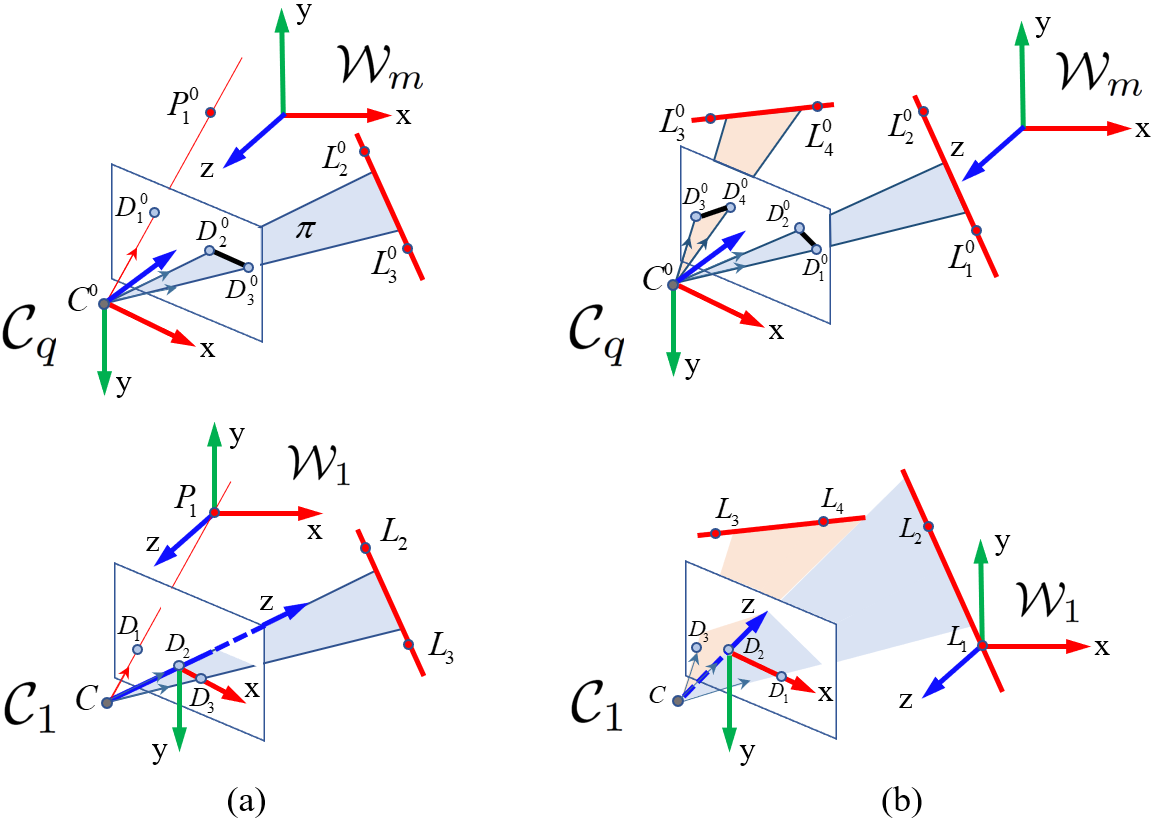}
    \caption{The illustration of intermediate reference frame $\mathcal{C}_{1}$ and $\mathcal{W}_{1}$ for (a) 1 point and 1 line case, (b) 2 lines case.}
    \label{fig.frames}
\end{figure}

To solve the 4DoF localization problem, we refer to two geometric constraints: collinearity of 3D-2D point matches and coplanarity of 3D-2D line matches. In an image, each pixel encodes a 3D projection ray. As shown in {Fig. \ref{fig.frames}}, according to the projection geometry, the optical center $C^{0}$ of the camera, the map point $P_{1}^{0}$ and its projection point $D_{1}^{0}$ on the image lie on the same 3D line, which is denoted as $\{C^{0},D_{1}^{0},{R_{{\mathcal{C}_{q}}{\mathcal{W}_{m}}}}P_{1}^{0}+{t_{{\mathcal{C}_{q}}{\mathcal{W}_{m}}}}\}_{L}$. By solving the line equation with the first two points $C^{0}$ and $D_{1}^{0}$, the third point can be substituted into the line equation to form two independent constraints. In addition, the camera center $C^{0}$, the map line $L_{2}^{0}L_{3}^{0}$ and its projection line segment $D_{2}^{0}D_{3}^{0}$ lie on the single plane $\pi$, which is denoted as ${\{C^{0},{D_{2}^{0}},{D_{3}^{0}},{R_{{\mathcal{C}_{q}}{\mathcal{W}_{m}}}}{L_{2}^{0}}+{t_{{\mathcal{C}_{q}}{\mathcal{W}_{m}}}}\}_{P}}$, and ${\{C,{D_{2}^{0}},{D_{3}^{0}},{R_{{\mathcal{C}_{q}}{\mathcal{W}_{m}}}}{L_{3}^{0}}+{t_{{\mathcal{C}_{q}}{\mathcal{W}_{m}}}}\}_{P}}$. Similarly, by solving the plane equation from the first three points $C^{0}$, $D_{2}^{0}$ and $D_{3}^{0}$, the two end points of the map line can be substituted into it to form another two independent constraints. As a result, two non-degenerate feature matches should be sufficient to solve the 4DoF localization problem with three possible combinations: 1 point 1 line, 2 points and 2 lines in theory.

%   \begin{figure}[tbp]
%        \centering
%        \includegraphics[width=0.35\textwidth]{fig/constraints}
%        \caption{ The collinearity and coplanarity constraints.}
%        \label{fig.constraints}
%    \end{figure}

\subsection{1 point 1 line}

In this section, we are going to solve the 4DoF localization problem given one 3D-2D point match and one 3D-2D line match. The solution is denoted as {{1P1L}}. Given the 2D coordinates of the detected image feature point, we can compute the corresponding 3D projection ray according to the calibrated intrinsic parameters. The same process is applied to the two end points of the line segment to obtain two 3D projection rays. The 3D projection rays are expressed in the camera reference frame $\mathcal{C}_{q}$, and the corresponding 3D coordinates of the features are expressed in the world reference frame $\mathcal{W}_{m}$. To simplify the form of the unsolved equations, we introduce two intermediate reference frames denoted as $\mathcal{C}_{1}$ and $\mathcal{W}_{1}$ for camera and map respectively.

\textbf{\emph{The choice of $\mathcal{C}_{1}$}}: As shown in {Fig. \ref{fig.frames}} (a), in the original camera reference frame $\mathcal{C}_{q}$, the origin is the camera center $C^{0}$, the camera projection ray associated with the 2D point is given by its normalized direction vector $\overrightarrow{d_{1}}$, and projection rays of the two end points of the 2D line are given by normalized direction $\overrightarrow{d_{2}}$, $\overrightarrow{d_{3}}$.

In the intermediate camera frame $\mathcal{C}_{1}$, the projection ray associated with the 2D point is denoted by $CD_{1}$, and the line, $CD_{2}$ and $CD_{3}$. Specifically, the intermediate reference frame $\mathcal{C}_{1}$ should satisfy the following conditions:
\begin{itemize}
\item The new camera center $C$ is $(0,0,-1)$.
\item One of the projection rays of the line end point $CD_{2}$ lies on the $z$ axis such that $D_{2}=(0,0,0)$.
\item The other projection ray of the line end point $CD_{3}$ lies on the $xz$ plane and the point $D_{3}$ is the intersection point between the $x$ axis and the ray.
\item The point $D_{1}$ in the projection ray of the point feature lies on the $xy$ plane.
\end{itemize}

After the transformation, we have following results
$$
C=\begin{bmatrix}
    0 \\
    0 \\
    -1
  \end{bmatrix},
D_{2}=\begin{bmatrix}
        0 \\
        0 \\
        0
      \end{bmatrix},
D_{3}=\begin{bmatrix}
  \tan(\arccos(\overrightarrow{d_{2}}\cdot\overrightarrow{d_{3}})) \\
  0 \\
  0
\end{bmatrix}
$$

Then the points in the original camera frame are calculated as follows
$$C^{0}={\bf{0}}_{3\times1}{\,},{\,}D_{2}^{0}=C^{0}+\overrightarrow{d_{2}}{\,},{\,}D_{3}^{0}=C^{0}+\frac{\overrightarrow{d_{3}}}{\overrightarrow{d_{2}}\cdot\overrightarrow{d_{3}}}$$

The transformation $T_{{\mathcal{C}_{1}}{\mathcal{C}_{q}}}$ can be computed by transforming the three points $(C^{0},D^{0}_{2},D^{0}_{3})$ to $(C,D_{2},D_{3})$. After that, the point $D_{1}\triangleq(a_{1},b_{1},0)$ can also be computed.

\textbf{\emph{The choice of $\mathcal{W}_{1}$}}: The transformation of the world reference is a translation which transforms the 3D point to the origin of $\mathcal{W}_{1}$
$$
T_{{\mathcal{W}_{1}}{\mathcal{W}_{m}}} =
\begin{bmatrix}
  {\bf{I}}_{3\times3} & -{P_{1}^{0}} \\
  {\bf{0}}_{1\times3} & 1
\end{bmatrix}
$$

Thus in $\mathcal{W}_{1}$ we have:
$$
P_{1}={\bf{0}}_{3\times1}, {\,}
L_{i=\{2,3\}}\triangleq
{{\begin{bmatrix}
      X_{i} &
      Y_{i} &
      Z_{i}
    \end{bmatrix}}^{T}}
$$

Note that these points are all known.
% collinearity of 2D-3D point match and coplanarity of 2D-3D line match

\emph{\textbf{Pose estimation between $\mathcal{C}_{1}$ and $\mathcal{W}_{1}$}}: Let us denote the rotation matrix and the translation matrix to be solved as $R$ and $t$, that is $R \triangleq {R_{{\mathcal{C}_{1}}{{\mathcal{W}}_{1}}}}$, $t \triangleq {t_{{\mathcal{C}_{1}}{{\mathcal{W}}_{1}}}}$. According to the collinearity of ${\{C,D_{1},RP_{1}+t\}}_{L}$, the following equations are derived:
\begin{equation}\label{eq.sub2}
  a_{1}T_{2}-b_{1}T_{1}=0
\end{equation}
\begin{equation}\label{eq.sub3}
  b_{1}T_{3}-T_{2}=-b_{1}
\end{equation}

As for the coplanarity of ${\{C,D_{2},D_{3},RL_{2}+t\}}_{P}$:
\begin{equation}\label{eq.sub4}
  R_{21}X_{2}+R_{22}Y_{2}+R_{23}Z_{2}+T_{2}=0
\end{equation}

And the coplanarity of ${\{C,D_{2},D_{3},RL_{3}+t\}}_{P}$:
\begin{equation}\label{eq.sub5}
  R_{21}X_{3}+R_{22}Y_{3}+R_{23}Z_{3}+T_{2}=0
\end{equation}
where $R_{mn}$ denotes the $m$-th row and $n$-th column entry of $R$, $T_{m}$ denotes the $m$-th entry of $t$.

One should notice that the matrix $R$ is determined only by $\alpha$. Combining (\ref{eq.sub4}) and (\ref{eq.sub5}), we can solve $\alpha$, which is then substituted in (\ref{eq.sub2}) - (\ref{eq.sub5}) for the translation $t$. Put all the equations together, the closed form localization problem can be solved as
$${T_{{\mathcal{C}_{q}}{\mathcal{W}_{m}}}}={{T_{{\mathcal{C}_{1}}{\mathcal{C}_{q}}}}^{-1}}\cdot{T_{{\mathcal{C}_{1}}{\mathcal{W}_{1}}}}\cdot{T_{{\mathcal{W}_{1}}{\mathcal{W}_{m}}}}$$

\emph{\textbf{Degenerate cases}}: For 1 point 1 line case, if the point lies on the line, the corresponding case is degenerated.

\subsection{2 points}

In this section, we are given two 3D-2D point matches, and the solution is denoted as {{2P}}. In this case, the intermediate camera reference frame remains the same as the original one, thus the points in $\mathcal{C}_{q}$ and $\mathcal{C}_{1}$ following the notations in 1P1L are
$$
C={\bf{0}}_{3\times1}, {\,}
D_{i=\{1,2\}}\triangleq
{{\begin{bmatrix}
      a_{i} &
      b_{i} &
      1
    \end{bmatrix}}^{T}}
$$
where $a_{1},a_{2},b_{1},b_{2}$ are known parameters computed using intrinsic parameters and normalized depth.

The transformation of the world reference is a translation which transforms one of the two 3D points to the origin of the intermediate world frame. Thus in $\mathcal{W}_{1}$ we have:
$$
P_{1}={\bf{0}}_{3\times1}, {\,}
P_{2}\triangleq
{{\begin{bmatrix}
      X_{2} &
      Y_{2} &
      Z_{2}
    \end{bmatrix}}^{T}}
$$

\emph{\textbf{Pose estimation between $\mathcal{C}_{1}$ and $\mathcal{W}_{1}$}}: Following the notations in 1P1L,  according to the collinearity of $\{C,D_{1},RP_{1}+t\}_L$, the following equations are derived:
\begin{equation}\label{eq.sub6}
  a_{1}T_{2}-b_{1}T_{1}=0
\end{equation}
\begin{equation}\label{eq.sub7}
  a_{1}T_{3}-T_{1}=0
\end{equation}

As for the collinearity of $\{C,D_{2},RP_{2}+t\}_L$, we have:
\begin{equation}\label{eq.sub8}
    \begin{split}
    &a_{2}(R_{21}X_{2}+R_{22}Y_{2}+R_{23}Z_{2}+T_{2}) \\
    &-b_{2}(R_{11}X_{2}+R_{12}Y_{2}+R_{13}Z_{2}+T_{1})=0
    \end{split}
\end{equation}
\begin{equation}\label{eq.sub9}
    \begin{split}
    &a_{2}(R_{31}X_{2}+R_{32}Y_{2}+R_{33}Z_{2}+T_{3}) \\
    &-(R_{11}X_{2}+R_{12}Y_{2}+R_{13}Z_{2}+T_{1})=0
    \end{split}
\end{equation}

Combining (\ref{eq.sub6}) - (\ref{eq.sub9}), we can solve ${T_{{\mathcal{C}_{1}}{{\mathcal{W}}_{1}}}}$. Thus the localization result can be obtained by
$${T_{{\mathcal{C}_{q}}{\mathcal{W}_{m}}}}={T_{{\mathcal{C}_{1}}{\mathcal{W}_{1}}}}\cdot{T_{{\mathcal{W}_{1}}{\mathcal{W}_{m}}}}$$

\subsection{2 lines}

In this section, we are given two 3D-2D line matches as shown in {Fig. \ref{fig.frames}} (b).

\textbf{\emph{The choice of $\mathcal{C}_{1}$}}: In $\mathcal{C}_{q}$, the camera projection rays associated with the two 2D lines are given by pairs $(\overrightarrow{d_{1}}, \overrightarrow{d_{2}})$ and $(\overrightarrow{d_{3}}, \overrightarrow{d_{4}})$, respectively. In $\mathcal{C}_{1}$, the 3D projection rays associated with the two 2D lines are represented by $(CD_{1},CD_{2})$, and $(CD_{2},CD_{3})$. Specifically, $\mathcal{C}_{1}$ should satisfy the following conditions:
\begin{itemize}

\item The new camera center $C$ is $(0,0,-1)$.
\item The intersection line of the two interpretation planes represented by projection ray $CD_{2}$ lies on the $z$ axis such that $D_{2}=(0,0,0)$.
\item The projection ray $CD_{1}$ lies on the $xz$ plane and the point $D_{1}$ is the intersection point between the $x$ axis and the ray.
\item The point $D_{3}$ in the projection ray of one line lies on the $xy$ plane.

\end{itemize}

The unit normal vectors of the two planes formed by $(C^{0},\overrightarrow{d_{1}}, \overrightarrow{d_{2}})$ and $(C^{0},\overrightarrow{d_{3}}, \overrightarrow{d_{4}})$ can be computed as follows
$$\overrightarrow{n_{1}}=\overrightarrow{d_{1}}\times\overrightarrow{d_{2}},
\overrightarrow{n_{2}}=\overrightarrow{d_{3}}\times\overrightarrow{d_{4}},
\overrightarrow{d_{12}}=\overrightarrow{n_{1}}\times\overrightarrow{n_{2}}
$$
where $\overrightarrow{d_{12}}$ is the direction vector of the intersection line $CD_{2}$. After such a transformation, we have the coordinates of $C$ and $D_{2}$, and $D_{1}$ can be computed as follows
$$
D_{1}={{\begin{bmatrix}
      \tan(\arccos(\overrightarrow{d_{1}}\cdot\overrightarrow{d_{12}})) &
      0 &
      0
    \end{bmatrix}}^{T}}
$$

Then the corresponding points in $\mathcal{C}_{q}$:
$$C^{0}={\bf{0}}_{3\times1}{\,},{\,}D_{1}^{0}=C^{0}+\frac{\overrightarrow{d_{1}}}{\overrightarrow{d_{1}}\cdot\overrightarrow{d_{12}}}{\,},{\,}D_{2}^{0}=C^{0}+\overrightarrow{d_{12}}$$

The transformation $T_{{\mathcal{C}_{1}}{\mathcal{C}_{q}}}$ can be computed by transforming $(C^{0},D^{0}_{1},D^{0}_{2})$ to $(C,D_{1},D_{2})$. After that, the point $D_{3}\triangleq(a_{1},b_{1},0)$ can also be computed.

\textbf{\emph{The choice of $\mathcal{W}_{1}$}}: $T_{{\mathcal{W}_{1}}{\mathcal{W}_{m}}}$ is a translation which transforms one end point of the 3D line to the origin of the intermediate world frame. Thus in $\mathcal{W}_{1}$ we have
$$
L_{1}={\bf{0}}_{3\times1}, {\,}
L_{\{i=2,3,4\}}\triangleq
{{\begin{bmatrix}
      X_{i} &
      Y_{i} &
      Z_{i}
    \end{bmatrix}}^{T}}
$$

\emph{\textbf{Pose estimation between $\mathcal{C}_{1}$ and $\mathcal{W}_{1}$}}: Following the notations in 1P1L, according to the coplanarity of $\{C,D_{1},D_{2},RL_{1}+t\}_P$ and $\{C,D_{1},D_{2},RL_{2}+t\}_P$, the following equations are derived
\begin{equation}\label{eq.sub10}
  T_{2}=0
\end{equation}
\begin{equation}\label{eq.sub11}
  R_{21}X_{2}+R_{22}Y_{2}+R_{23}Z_{2}+T_{2}=0
\end{equation}

As for the coplanarity of ${\{C,D_{2},D_{3},RL_{3}+t\}}_{P}$ and ${\{C,D_{2},D_{3},RL_{4}+t\}}_{P}$:
\begin{equation}\label{eq.sub12}
    \begin{split}
    &a_{1}(R_{21}X_{3}+R_{22}Y_{3}+R_{23}Z_{3}+T_{2}) \\
    &-b_{1}(R_{11}X_{3}+R_{12}Y_{3}+R_{13}Z_{3}+T_{1})=0
    \end{split}
\end{equation}
\begin{equation}\label{eq.sub13}
    \begin{split}
    &a_{1}(R_{21}X_{4}+R_{22}Y_{4}+R_{23}Z_{4}+T_{2}) \\
    &-b_{1}(R_{11}X_{4}+R_{12}Y_{4}+R_{13}Z_{4}+T_{1})=0
    \end{split}
\end{equation}

From (\ref{eq.sub10}) - (\ref{eq.sub13}), we can easily find that the constraints are not sufficient to solve $T_{3}$. In fact, one of the constraints provided by IMU is coincident with one of constraints provided by coplanarity, thus the 2 lines case cannot solve the 4DoF localization problem.

\section{Model Selection}

\begin{figure*}[tbp]
    \centering
    \subfigure[10 matches]{
        \includegraphics[width=0.22\textwidth]{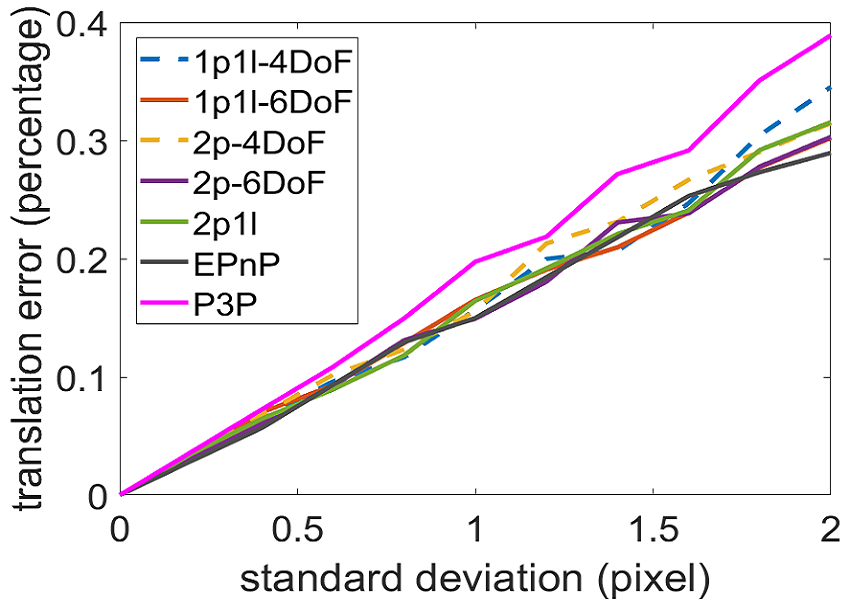}
        \includegraphics[width=0.22\textwidth]{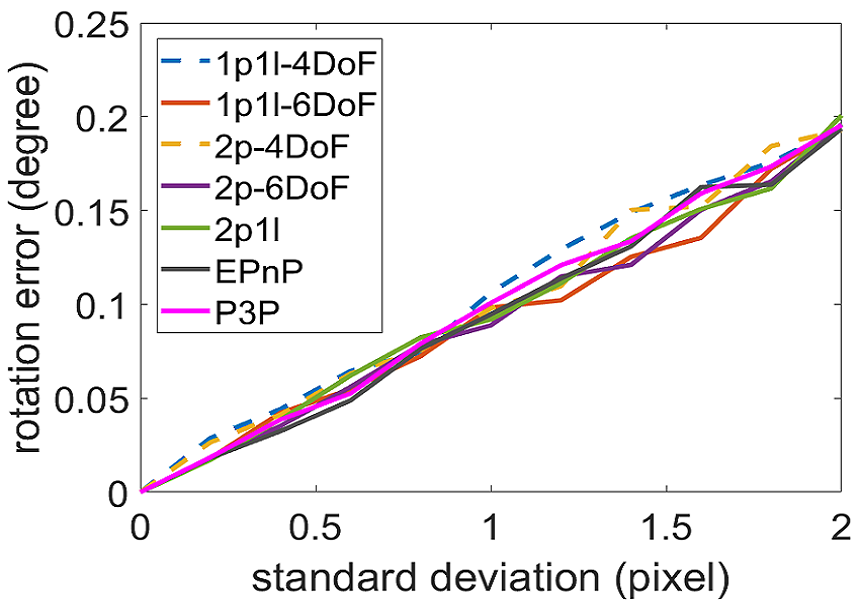}
    }
    \subfigure[5 matches]{
        \includegraphics[width=0.22\textwidth]{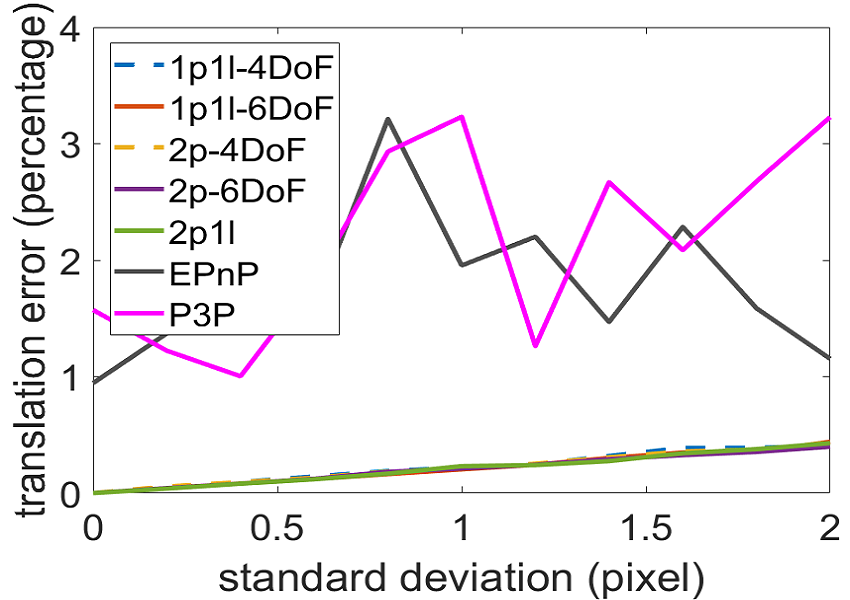}
        \includegraphics[width=0.22\textwidth]{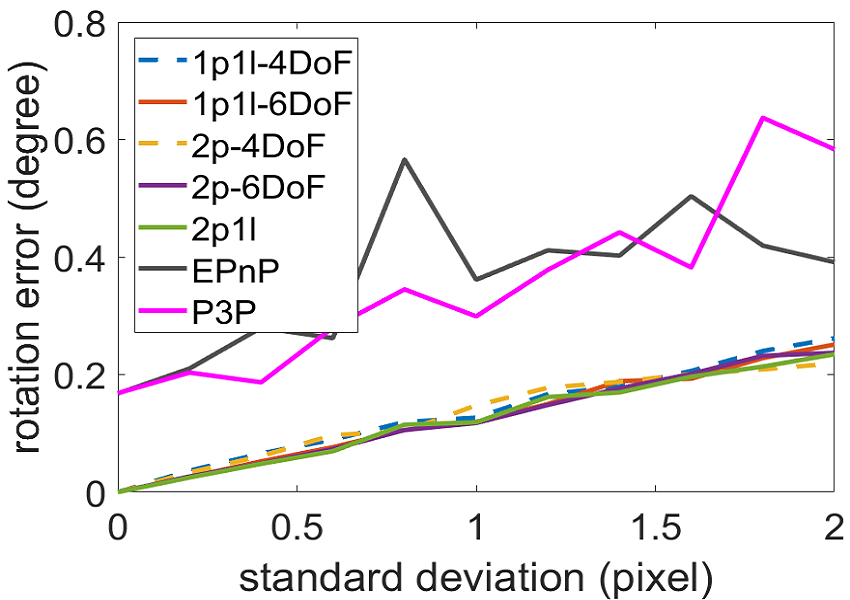}
    }
    \quad
    \subfigure[4 matches]{
        \includegraphics[width=0.22\textwidth]{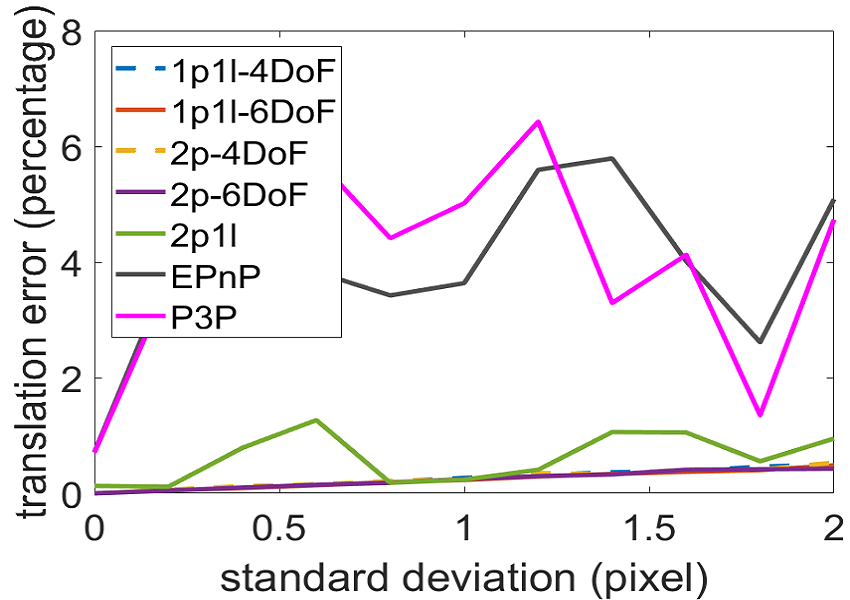}
        \includegraphics[width=0.22\textwidth]{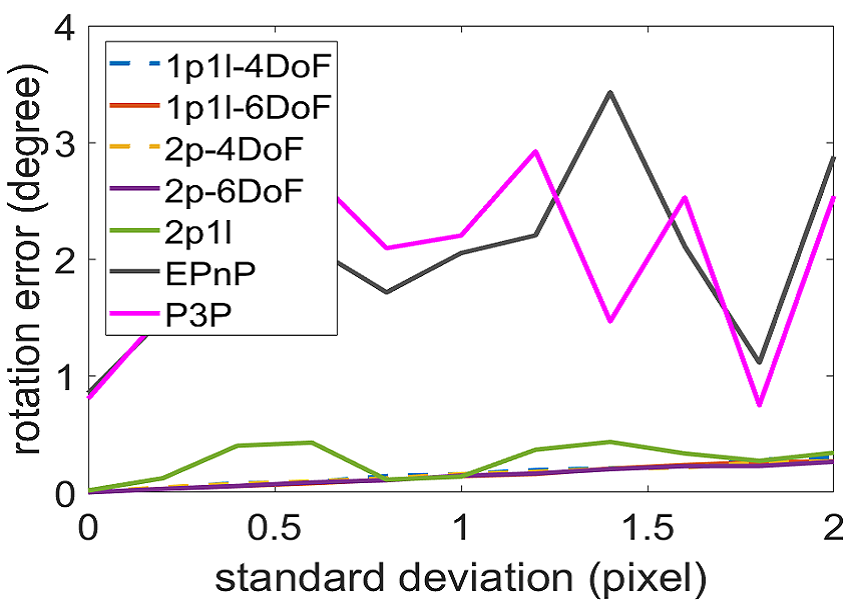}
    }
    \subfigure[3 matches]{
        \includegraphics[width=0.22\textwidth]{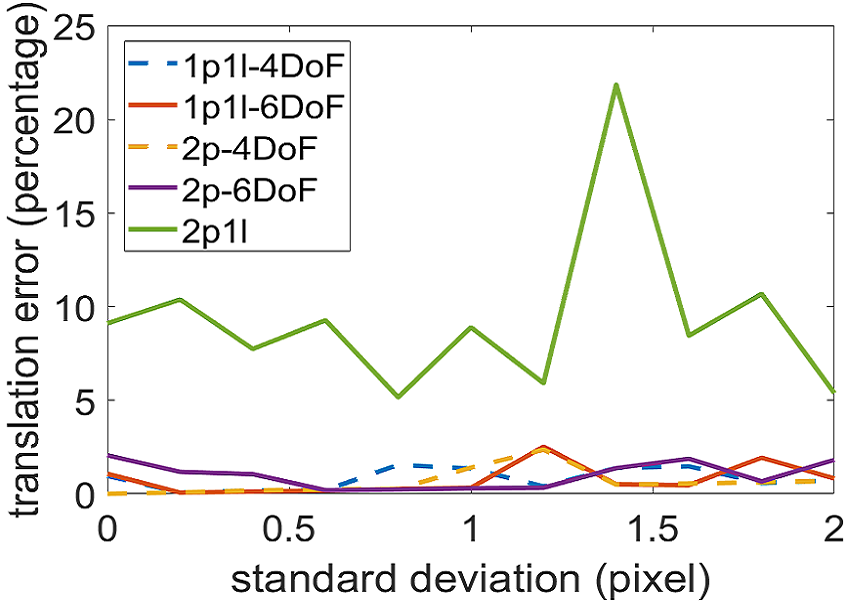}
        \includegraphics[width=0.22\textwidth]{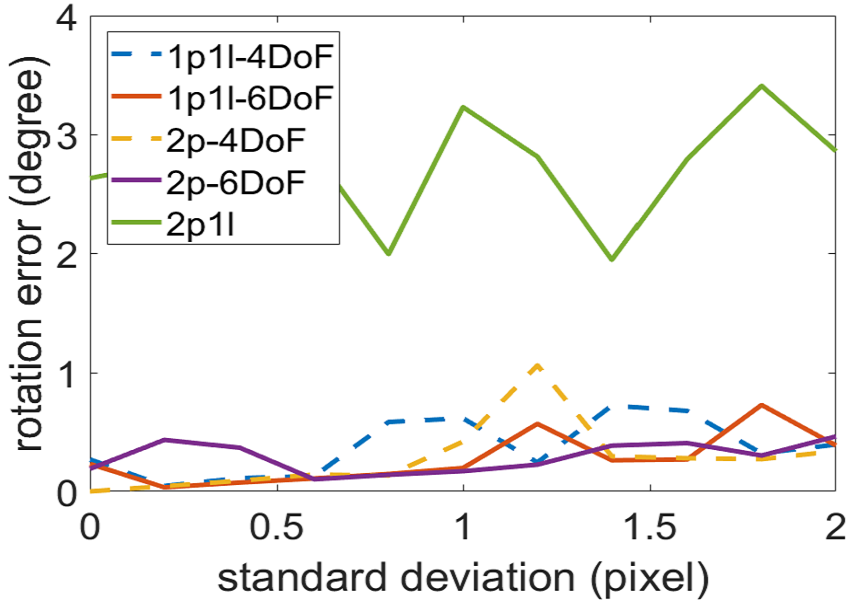}
    }
    \caption{Accuracy simulation results: the translation and rotation error for 10, 5, 4, 3 feature matches of different methods. Notice that EPnP and P3P are not computed in 3 matches case, for they all need at least four point matches to compute.}
    \label{fig.accuracy}
\end{figure*}

In the last section, we discussed the possible minimal closed form solutions of 4DoF localization problem when inertial measurements exist, and showed that the {{1P1L}} and {{2P}} are solvable. Based on the two minimal solutions, we propose the 2-entity RANSAC, in which there are three possible sampling strategies according to the corresponding solution: {{1P1L}}, {{2P}} and {{mixed}}. The {{mixed}} refers to using both {{1P1L}} and {{2P}} solutions according to the randomly selected features. Specifically, we first select one point match and then randomly select another feature match from points and lines. If it is a point match, we perform {{2P}}, if it is a line match, we perform {{1P1L}}. There are some works discuss about sampling points from multiple combinations of views \cite{clipp2009new} \cite{vasconcelos2017automatic}, while we are focusing on sampling from multiple types of features (points and lines). In this section, we derive the success probability of the three sampling strategies to provide insights of sampling in 2-entity RANSAC.

Let's denote $p$ as the number of point matches, $l$ as the number of line matches, $\lambda$ as the point inliers rate, and $\gamma$ as the line inliers rate.
$$
\frac{m}{p}=\lambda , \frac{n}{l}=\gamma {\,} (0\leq \lambda,\gamma \leq 1) ,
\frac{l}{p}=\varepsilon {\,} (0\leq\varepsilon\leq l)
$$
where $m$, $n$ denote the point inliers number and line inliers number respectively. Then the success probability of different sampling strategies during one iteration in RANSAC is derived as follows:
\begin{align}
  P_{1p1l} &= \frac{m}{p} \cdot \frac{n}{l} = \lambda\cdot\gamma \nonumber\\
  P_{2p} &= \frac{m}{p} \cdot \frac{m-1}{p-1} = \lambda \cdot \frac{\lambda p - 1}{p-1} \nonumber\\
  P_{mixed} &= \frac{m}{p} \cdot \frac{m+n-1}{p+l-1} = \lambda\cdot\frac{\lambda p+\gamma l-1}{p+l-1} \nonumber
\end{align}

Then we have
%$$
%P_{1p1l}-P_{mixed} = \lambda(\gamma-\frac{\lambda p+\gamma l-1}{p+l-1}) \varpropto \gamma-(\lambda-a)
%$$
%$$
%P_{mixed}-P_{2p} = \lambda(\frac{\lambda p+\gamma l-1}{p+l-1}-\frac{\lambda p-1}{p-1}) \varpropto \gamma-(\lambda-a)
%$$
\begin{align}
  &P_{1p1l}-P_{mixed} = \lambda(\gamma-\frac{\lambda p+\gamma l-1}{p+l-1}) \varpropto \gamma-(\lambda-a) \nonumber \\
  &P_{mixed}-P_{2p} = \lambda(\frac{\lambda p+\gamma l-1}{p+l-1}-\frac{\lambda p-1}{p-1}) \varpropto \gamma-(\lambda-a) \nonumber
\end{align}
where $a = \frac{1-\lambda}{p-1}$.\\

Generally, $p-1\gg1-\lambda$ holds for most real world applications, which means $a$ is a small positive number close to 0. Thus, the following conclusion can be derived:
\begin{align}
  \gamma \geq \lambda \Rightarrow P_{1p1l} > P_{mixed} > P_{2p} \nonumber
\end{align}
%$$
%\gamma \geq \lambda \Rightarrow P_{1p1l} > P_{mixed} > P_{2p}
%$$
Furthermore, with proper scaling, the following conclusion can also hold:
\begin{align}
  \gamma < \lambda \Rightarrow P_{1p1l} \leq P_{mixed} \leq P_{2p} \nonumber
\end{align}
%$$
%\gamma < \lambda \Rightarrow P_{1p1l} \leq P_{mixed} \leq P_{2p}
%$$

The percentage of line features and point features is highly related to the environment. According to the above discussion, we could find that the {{mixed}} sampling strategy is a more robust selection to the environmental variations as its performance is always the average one. Empirically, when localizing in the unstructured scene, there might be more point feature matches which indicates a  higher point inlier rate than line features, leading to the choice of 2P sampling strategy. In the structured scene, although there are more line features compared with the unstructured scene, there are also many good point features. Considering the reconstruction error of the 3D lines can be higher than that of 3D points, the performance of 1P1L may not be significantly better than the 2P method in the structured scene. In order to be adaptive to the changing environment, an automatic selection mechanism is proposed by statistic estimation of the inlier rate using historical data. Specifically, the map is divided into structured or unstructured segments according to the inlier rate of point and line features. Then, when the robot is in the segment, the corresponding sampling strategy is selected, e.g. 2P in unstructured environment.

\begin{figure*}[htbp]
    \centering
    \includegraphics[width=\textwidth]{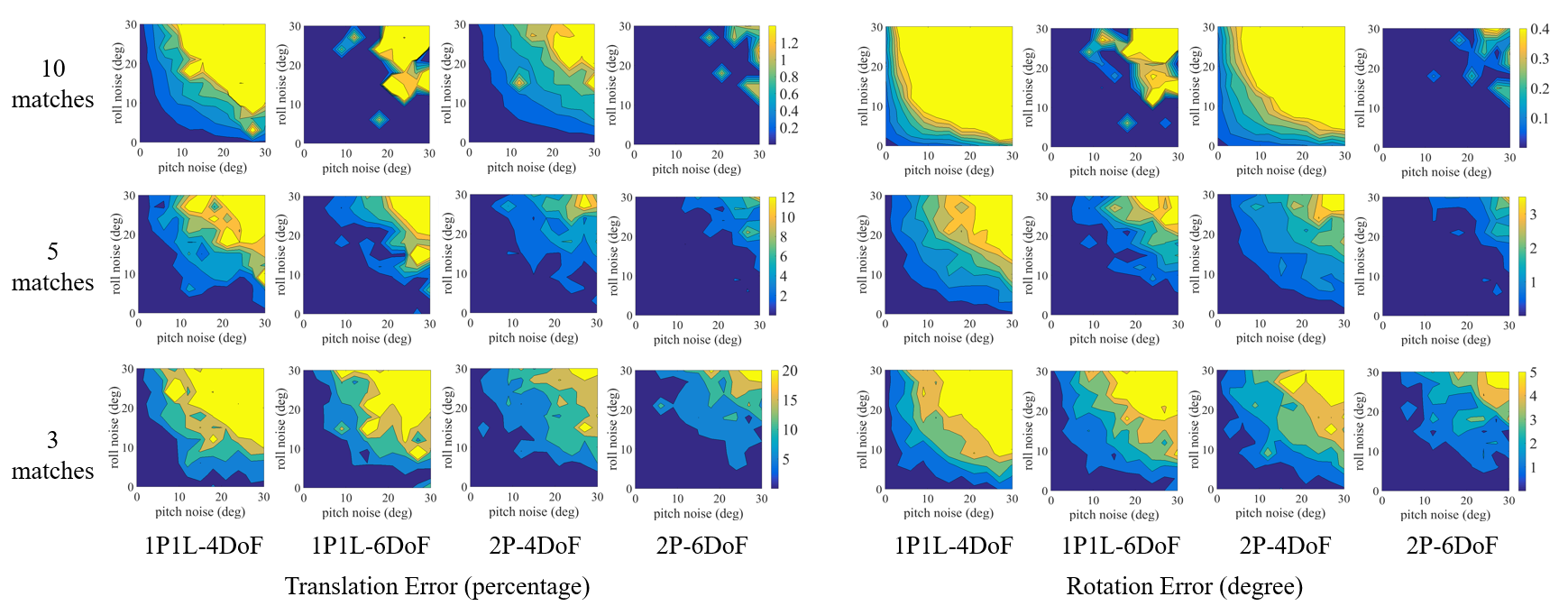}
    \caption{Sensitivity simulation results: the translation and rotation error for 10, 5, 3 feature matches, respectively.}
    \label{fig.sensitivity}
\end{figure*}

\section{Experimental Results}

The proposed methods are evaluated on both synthetic and real data to demonstrate the advantages over other baseline methods in visual localization. We start by simulation experiments to illustrate the accuracy when the image feature is noisy, the sensitivity with respect to inaccurate pitch and roll angle, and the robustness against outliers. Then on real world data, we demonstrate the success rate comparison to show the effectiveness of the proposed methods and the model selection mechanism.

\subsection{Results with synthetic data}

In the simulation experiments, we generated some 3D points and lines in the cube $[-1,1]^{3}$ and computed 2D projections for varying camera poses to get 3D-2D feature matches. For each method, 100 iterations of RANSAC are performed. The final identified inliers are sent to a nonlinear minimization for high accuracy. In optimization, there are two possible ways: 4DoF or 6DoF optimization. The 4DoF optimization means we fix the pitch and roll angle provided by IMU, and only optimize the other four variables in the pose. This may be useful when there are little inliers to optimize.

The minimal solutions evaluated in simulation experiments are EPnP\cite{lepetit2009epnp}, P3P\cite{gao2003complete}, 2P1L\cite{ramalingam2011pose} and our proposed solutions including 1P1L-6DoF, 1P1L-4DoF, 2P-6DoF, 2P-4DoF. The mixed-6DoF and mixed-4DoF are also evaluated in the robustness against outliers experiment. We exploited the translation and rotation error to evaluate the estimated result $[R|t]$ with respect to the ground truth $[R_{gt}|t_{gt}]$. Following \cite{ramalingam2011pose}, for translation error, we computed $\|t-t_{gt}\|/t_{gt}$ in percentage. For rotation error, we first computed the relative rotation $\triangle R=RR_{gt}^{T}$ and represented it in degrees.

\subsubsection{Accuracy}

To quantify the accuracy of different minimal solutions, we added Gaussian noise with zero mean and varying standard deviations for the 2D projections and changed the number of feature matches in four levels: 10, 5, 4, 3 (the 10 case means there are 10 point matches and 10 line matches in the scene). The results are shown in {Fig. \ref{fig.accuracy}}. We could find that when the feature matches are sufficient (see the 10 case), the proposed methods can achieve the same accuracy over other baseline methods as the standard deviation of the noise increases. While the number of feature matches decreases, the advantage of our 2-entity methods becomes bigger compared with 2P1L, EPnP and P3P. One can notice that when the number of feature matches decreases to 4, the error of our methods is slightly larger than that of the 10 matches case, while others grow severely.

\subsubsection{Sensitivity}

With the inertial sensor's aid, we reduce the degree of the localization problem to 4 by leveraging the pitch and roll angle provided by the sensor. Therefore, it's necessary to investigate the influence of the quality of the two angles on the final accuracy. We added Gaussian noise with zero mean and varying standard deviations on the pitch and roll angle and studied the performance of the proposed methods in three feature matches levels: 10, 5, 3. The results are shown in {Fig. \ref{fig.sensitivity}}.

When there are enough feature matches (see the 10 case), the proposed methods can tolerate almost 25 degrees of noise on both pitch and roll angles. Empirically in real world application, the noise on pitch and roll angle is far less, thus the noise on pitch and roll should have limited impact on the accuracy. We also note that the 6DoF optimization outperforms the 4DoF in this case. However, as the number of feature matches decreases, the difference between the 6DoF and 4DoF methods becomes smaller. This is reasonable because when the inliers number is few, the error caused by the additional DoFs in optimization may be larger than that caused by noisy attitude estimation. Therefore, if there are very limited reliable feature matches when long term change occurs, 4DoF optimization is a good choice.

\begin{figure*}[htbp]
        \centering
        \subfigure[10 matches]{
            \includegraphics[width=0.22\textwidth]{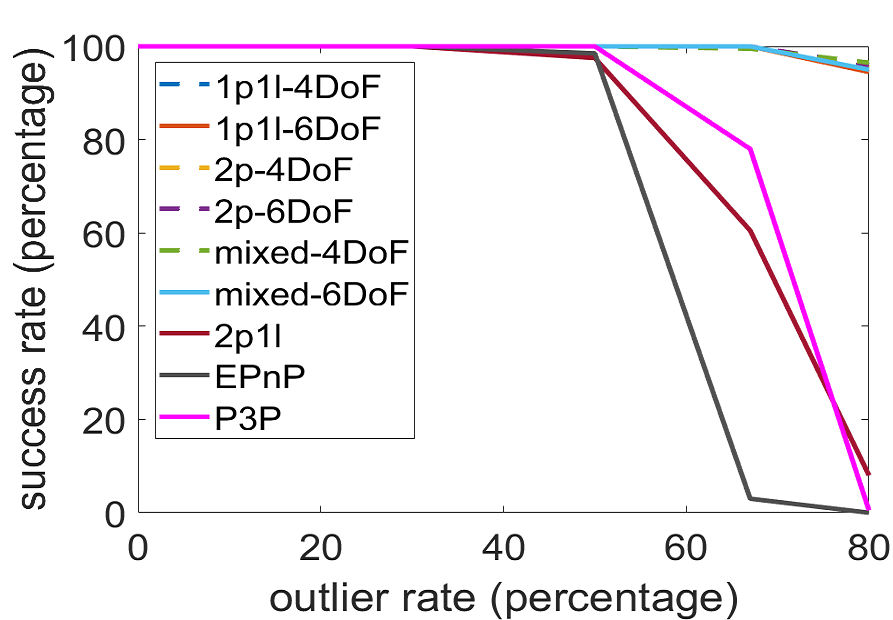}
        }
        \subfigure[5 matches]{
            \includegraphics[width=0.22\textwidth]{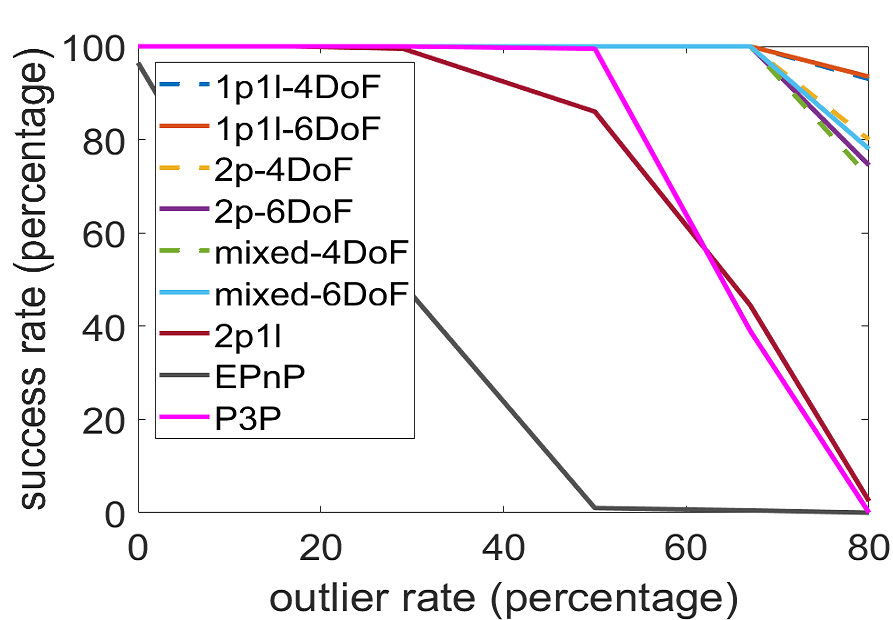}
        }
        \subfigure[4 matches]{
            \includegraphics[width=0.22\textwidth]{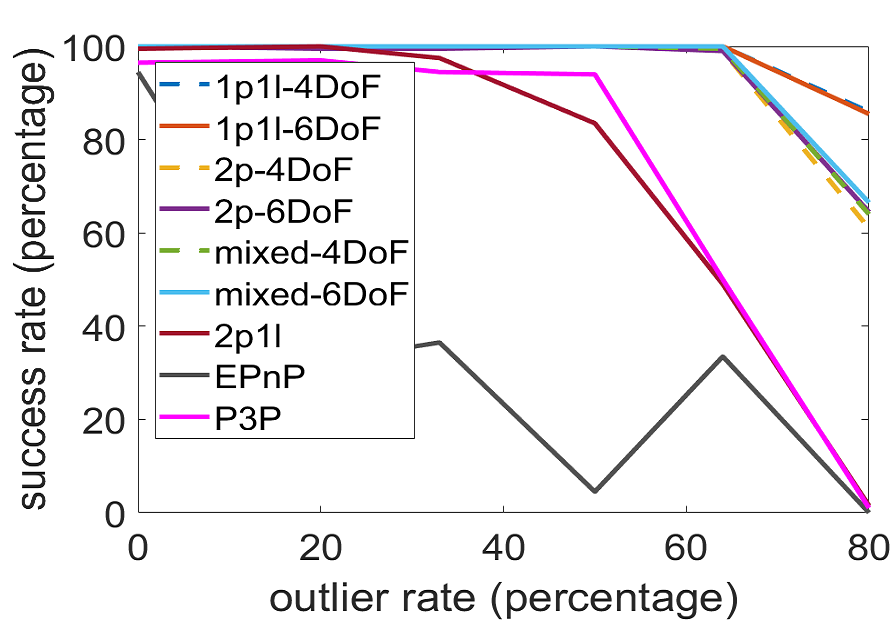}
        }
        \subfigure[3 matches]{
            \includegraphics[width=0.22\textwidth]{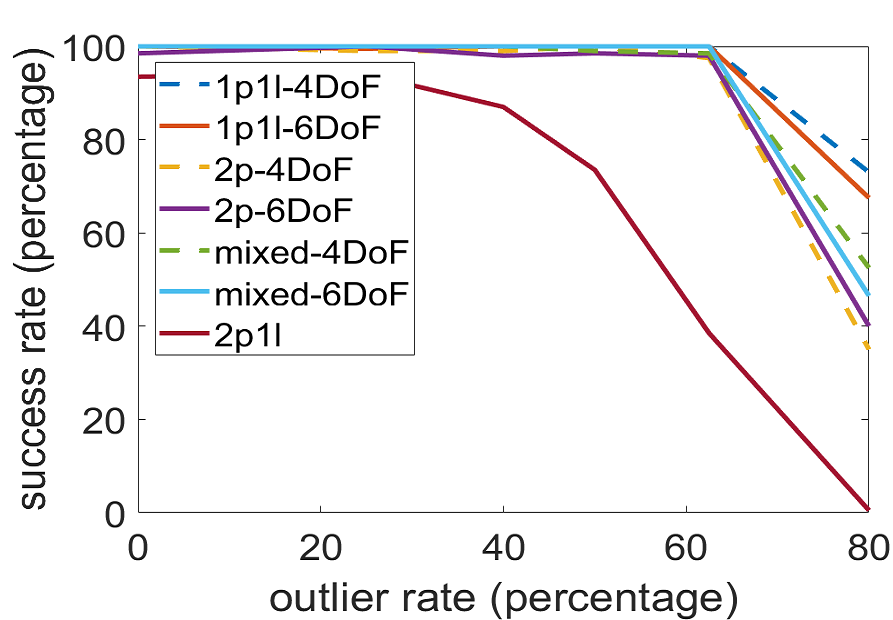}
        }
        \caption{Robustness simulation results: the success rate for 10, 5, 4, 3 inliers case of different methods.}
        \label{fig.robustness}
    \end{figure*}

\begin{table*}[tbp]
\caption{Evaluation of the localization on the selected scene of YQ-Dataset.}
\begin{center}
\resizebox{\textwidth}{!}{
\begin{tabular}{c|c|c||c|c||c|c}
\multirow{2}{*}{} & \multicolumn{2}{c||}{2017-0827} & \multicolumn{2}{c||}{2017-0828} & \multicolumn{2}{c}{2018-0129} \\
 & unstructured scene & structured scene & unstructured scene & structured scene & unstructured scene & structured scene \\
\cline{2-7}
\makecell[r]{Translation Error (m) \\Rotation Error (deg)} & \makecell[c]{.25/.5/1.0/5.0 \\ 2/5/8/10} & \makecell[c]{.25/.5/1.0/5.0 \\ 2/5/8/10} & \makecell[c]{.25/.5/1.0/5.0 \\ 2/5/8/10} & \makecell[c]{.25/.5/1.0/5.0 \\ 2/5/8/10} & \makecell[c]{.25/.5/1.0/5.0 \\ 2/5/8/10} & \makecell[c]{.25/.5/1.0/5.0 \\ 2/5/8/10} \\
\hline
1P1L-4DoF & \textcolor{black}{0.0} / \textcolor{white}{0}{1.9} / \textcolor{white}{0}{3.8} / \textcolor{black}{15.0} &
\textcolor{white}{0}{7.9} / \textcolor{black}{23.7} / \textcolor{black}{42.1} / \textcolor{black}{81.6}  & \textcolor{white}{0}{1.9} / \textcolor{white}{0}{9.4} / \textcolor{black}{15.0} / \textcolor{black}{29.4} & \textcolor{black}{13.0} / \textcolor{black}{32.6} / \textcolor{black}{54.3} / \textcolor{black}{87.0} & \textcolor{black}{ 0.0} / \textcolor{black}{ 0.0} / \textcolor{black}{ 0.0} / \textcolor{black}{ 3.8} & \textcolor{white}{0}{5.0} / \textcolor{black}{15.5} / \textcolor{black}{35.5} / \textcolor{black}{85.0} \\
\hline
1P1L-6DoF & \textcolor{black}{1.9} / \textcolor{white}{0}{5.7} / \textcolor{white}{0}{8.8} / \textcolor{black}{18.8} & \textcolor{blue}{29.5} / \textcolor{black}{48.6} / \textcolor{black}{78.4} / \textcolor{red}{91.9} & \textcolor{white}{0}{6.3} / \textcolor{black}{13.1} / \textcolor{black}{21.9} / \textcolor{black}{35.0} & \textcolor{blue}{39.1} / \textcolor{black}{67.4} / \textcolor{black}{82.6} / \textcolor{black}{91.3} & \textcolor{black}{ 0.0} / \textcolor{black}{ 0.0} / \textcolor{black}{ 0.0} / \textcolor{black}{ 4.9} & \textcolor{red}{25.0} / \textcolor{blue}{45.0} / \textcolor{blue}{65.0} / \textcolor{blue}{90.0} \\
\hline
2P-4DoF & \textcolor{black}{ 2.5} / \textcolor{black}{11.3} / \textcolor{black}{18.8} / \textcolor{black}{36.3} & \textcolor{black}{20.5} / \textcolor{black}{31.8} / \textcolor{black}{50.0} / \textcolor{black}{70.5} & \textcolor{black}{11.3} / \textcolor{black}{29.4} / \textcolor{black}{42.5} / \textcolor{black}{56.3} & \textcolor{black}{21.7} / \textcolor{black}{52.2} / \textcolor{black}{67.4} / \textcolor{black}{93.5} & \textcolor{black}{ 0.0} / \textcolor{black}{ 0.0} / \textcolor{red}{ 1.0} / \textcolor{black}{ 6.8} & \textcolor{black}{10.0} / \textcolor{black}{20.0} / \textcolor{black}{30.0} / \textcolor{black}{80.0} \\
\hline
2P-6DoF & \textcolor{blue}{ 6.3} / \textcolor{red}{15.0} / \textcolor{blue}{24.4} / \textcolor{blue}{40.0} & \textcolor{red}{31.8} / \textcolor{blue}{57.9} / \textcolor{red}{83.9} / {86.8} & \textcolor{blue}{25.0} / \textcolor{red}{38.1} / \textcolor{red}{48.8} / \textcolor{red}{58.1} & \textcolor{red}{45.7} / \textcolor{blue}{69.6} / \textcolor{blue}{91.3} / \textcolor{blue}{95.7} & \textcolor{red}{1.0} / \textcolor{red}{1.0} / \textcolor{red}{1.0} / \textcolor{red}{9.7} & \textcolor{red}{25.0} / {40.0} / \textcolor{blue}{65.0} / \textcolor{red}{95.0} \\
\hline
mixed-4DoF & \textcolor{black}{4.4} / \textcolor{blue}{13.8} / \textcolor{red}{26.9} / \textcolor{red}{44.4} & \textcolor{black}{13.6} / \textcolor{black}{31.8} / \textcolor{black}{47.7} / \textcolor{black}{72.7} & \textcolor{black}{10.6} / \textcolor{black}{30.0} / \textcolor{black}{37.5} / \textcolor{blue}{56.9} & \textcolor{black}{21.7} / \textcolor{black}{43.5} / \textcolor{black}{63.0} / \textcolor{blue}{95.7} & \textcolor{black}{0.0} / \textcolor{black}{0.0} / \textcolor{red}{1.0} / \textcolor{blue}{7.8} & \textcolor{white}{0}{5.0} / \textcolor{black}{25.0} / \textcolor{black}{35.0} / \textcolor{black}{80.0} \\
\hline
mixed-6DoF & \textcolor{red}{ 6.9} / \textcolor{red}{15.0} / \textcolor{black}{23.4} / \textcolor{black}{38.8} & \textcolor{red}{31.8} / \textcolor{red}{63.2} / \textcolor{black}{78.9} / \textcolor{blue}{89.6} & \textcolor{red}{25.6} / \textcolor{blue}{36.9} / \textcolor{blue}{48.1} / \textcolor{red}{58.1} & \textcolor{red}{45.7} / \textcolor{red}{73.9} / \textcolor{red}{93.5} / \textcolor{red}{97.8} & \textcolor{red}{ 1.0} / \textcolor{red}{ 1.0} / \textcolor{red}{ 1.0} / \textcolor{blue}{7.8} & \textcolor{blue}{20.0} / \textcolor{red}{55.0} / \textcolor{red}{70.0} / \textcolor{red}{95.0} \\
\hline
P3P & \textcolor{black}{1.9} / \textcolor{black}{11.9} / \textcolor{black}{20.0} / \textcolor{black}{37.5} & \textcolor{blue}{29.5} / \textcolor{black}{50.0} / \textcolor{black}{73.7} / \textcolor{black}{86.8} & \textcolor{black}{16.3} / \textcolor{black}{26.9} / \textcolor{black}{39.4} / \textcolor{black}{52.5} & \textcolor{black}{30.4} / \textcolor{black}{63.0} / \textcolor{black}{76.1} / \textcolor{black}{91.3} & \textcolor{black}{ 0.0} / \textcolor{black}{ 0.0} / \textcolor{red}{ 1.0} / \textcolor{black}{2.9} & \textcolor{black}{10.0} / \textcolor{black}{40.0} / \textcolor{black}{60.0} / \textcolor{blue}{90.0} \\
\hline
EPnP & \textcolor{black}{4.4} / \textcolor{black}{11.9} / \textcolor{black}{20.6} / \textcolor{black}{35.7} & \textcolor{black}{27.3} / \textcolor{black}{55.3} / \textcolor{blue}{81.6} / \textcolor{black}{89.5} & \textcolor{black}{16.9} / \textcolor{black}{31.3} / \textcolor{black}{40.6} / \textcolor{black}{56.3} & \textcolor{black}{26.1} / \textcolor{black}{65.2} / \textcolor{black}{84.8} / \textcolor{black}{93.5} & \textcolor{black}{ 0.0} / \textcolor{black}{ 0.0} / \textcolor{red}{ 1.0} / \textcolor{black}{2.9} & \textcolor{blue}{20.0} / \textcolor{black}{40.0} / \textcolor{blue}{65.0} / \textcolor{blue}{90.0} \\
\end{tabular}}
\end{center}
\label{table.select}
\end{table*}

\subsubsection{Robustness}

To validate the robustness of the proposed methods, we designed a few experiments by varying the outlier rate from 0 to 80\% with different number levels of inliers: 10, 5, 4, 3. We achieve the outlier rate by adding outliers and the outlier is generated by incorrectly associating the features in the original data. When the translation error is less than 10\% and the rotation error is smaller than 5 degrees, the localization is assumed to be successful. We did 200 trails for each method to average the success rate. The results can be seen in {Fig. \ref{fig.robustness}}. As expected, the proposed 2-entity RANSAC outperforms the other minimal solutions when the outlier rate increases, which is more obvious when the number of inliers decreases. Note that with sufficient inliers, the proposed methods can achieve a success rate of more than 90\% when the outlier rate is 80\%.

\subsection{Results with real data}

For real world experiments, YQ-Dataset is utilized, and three of the datasets were collected during three days in summer 2017, denoted as 2017-0823, 2017-0827, 2017-0828, and the other was collected in winter 2018, denoted as 2018-0129. We select the 2017-0823 dataset to build the 3D map and utilize the other three datasets to evaluate the localization performance over different weathers or seasons. The ground truth relative pose is provided by aligning the synchronized 3D LiDAR scans. The evaluated methods are EPnP, P3P, and our methods including 1P1L-4/6DoF, 2P-4/6DoF, mixed-4/6DoF. To get the 3D-2D feature matches between the query image and the map, we exploited the following steps:
\begin{itemize}
\item Obtain the camera poses and the 3D-2D point matches in the map using visual inertial SLAM software \cite{mur2017visual}.
\item Run Line3D++ algorithm \cite{hofer2017efficient} to get the 3D-2D line matches in the map.
\item For the query session, we get the 3D-2D points/lines match based on the descriptors of LibVISO2 \cite{Geiger2011IV} and LBD \cite{zhang2013efficient}, which are fed to the RANSAC with different solutions.
\item For the pitch and roll angle of the query image, we directly use the corresponding values measured by IMU.
\end{itemize}

We first counted the correctly localized query images out of all keyframes in the query session using different methods, and represent the successful localization rate in percentage under four translation and rotation error thresholds, which can be seen in {Tab. \ref{table.whole}}. As expected, the performance of methods with 6DoF optimization are still better than 4DoF optimization. Besides, the 2-entity RANSAC is better than RANSAC with P3P and EPnP which calls for more number of matches. The proposed 2P and mixed are obviously better than EPnP and P3P, which indicates the robustness of the proposed methods when dealing with different weathers and seasons. Note that 1P1L is not as good as point features based methods, of which the reason is that there are many trees on both sides of the road in the map environment so that point features are far more abundant than line features.

\begin{figure}[tp]
  \centering
  \includegraphics[width=0.44\textwidth]{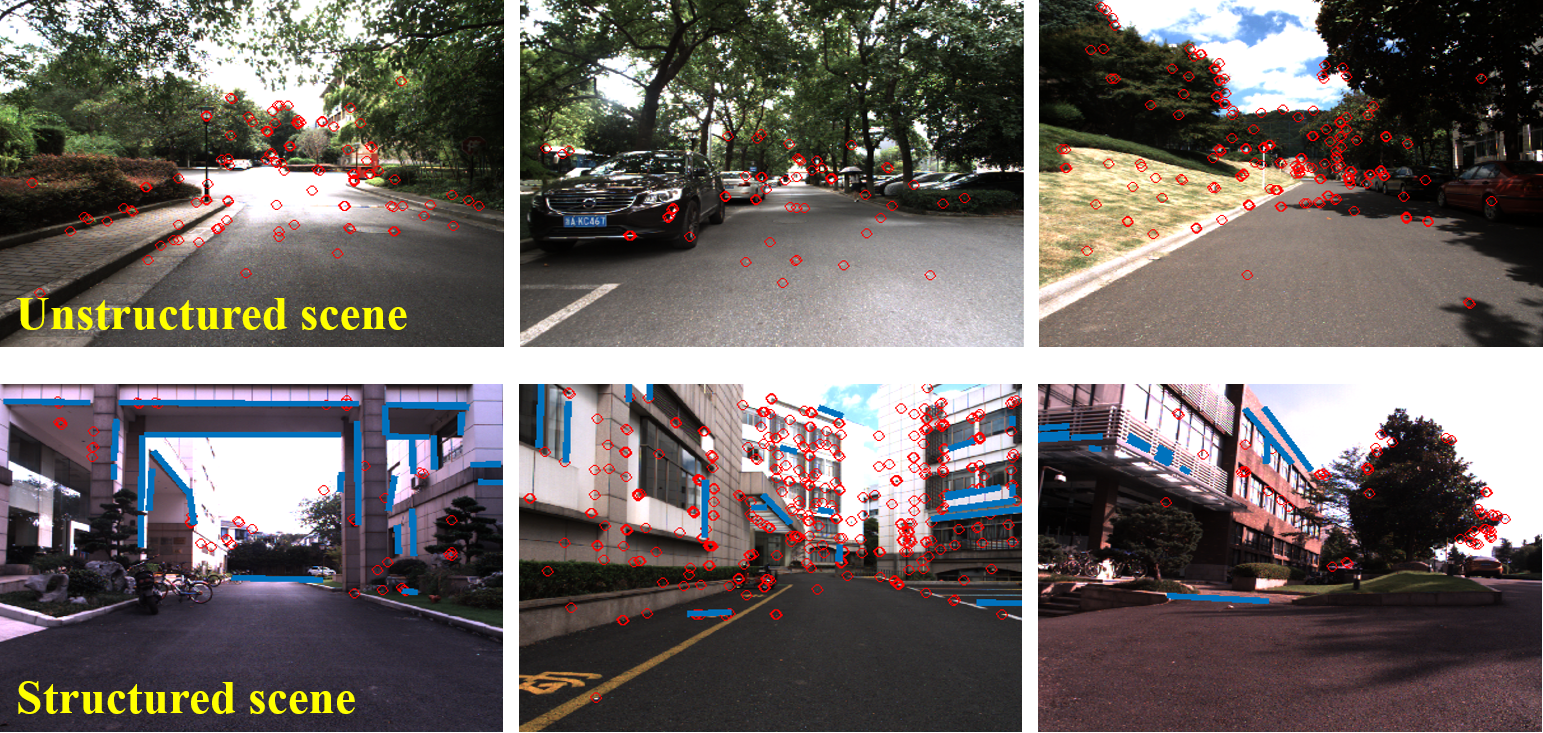}
  \caption{Examples of unstructured and structured scene.}
  \label{fig.scene}
\end{figure}

In order to make a fair comparison between the 1P1L and the 2P methods, we manually picked some structured and unstructured segments from the map, in which some images are shown in {Fig. \ref{fig.scene}}. There are around 100 places selected in unstructured segments, and 50 places in structured segments. The success rates of localization with multiple methods on the selected segments are show in {Tab. \ref{table.select}}. The results show that the 2P method performs best in the unstructured scene and the performance of 1P1L is similar with 2P in the structured scene but obviously worse than that of 2P in unstructured segments, confirming the fact that the feature distribution in the environment has unavoidable effect on the method selection. Besides, from the results of the structured scene in 2018-0129, we can find that when outlier percentage grows due to the changing season, the better robustness of line feature matches promotes the performance of 1P1L method. The mixed-6DoF, as derived in theory, gives relatively stable performance compared with 2P and 1P1L in all segments.

To further verify the correctness of the model selection mechanism, we did an extra leave-one-out experiment on the 2017-0828 dataset. According to the localization results of 2017-0827 and 2018-0129 dataset on the 2017-0823 map, we automatically labeled the whole dataset with structured and unstructured segments by the inlier rate of the point and line features acquired by the estimated pose using 2P-6DoF. If the inlier rate of lines was higher than the points, 1P1L is utilized in that segment, otherwise, 2P. After that, we counted the success rate on the whole 2017-0828 dataset again over different localization error thresholds and calculate the area under the curve of each method, which is shown in {Tab. \ref{table.selection}}. Results show that with the selection mechanism, the performance is further promoted as expected.

%\begin{figure}[htp]
%  \centering
%  \includegraphics[width=0.48\textwidth]{fig/scene}
%  \caption{The overview of the dataset trajectory. The 1, 3, 4 subsegments of the trajectory are representing the unstructured scene with more point features, and the 2, 5, 6 subsegments are the structured scene with more line features. The segments are labeled manually according to experience. }
%  \label{fig.dataset}
%\end{figure}

\section{Conclusions}

In this paper, the visual localization problem is reduced to 4DoF with the aid of inertial measurements. Based on this, the unified 2-entity RANSAC pipeline is developed and validated on both synthetic and real data to demonstrate the robustness and efficiency of the methods. Furthermore, we have investigated the proper mechanism to select the appropriate sampling strategy and showed the practicability in the real world application.

\begin{table}[tbp]
\caption{Evaluation on the whole YQ-Dataset.}
\begin{center}
\resizebox{0.5\textwidth}{!}{
\begin{tabular}{c|c||c||c}
%\multirow{2}{*}{} & \multicolumn{2}{c||}{2017-0827} & \multicolumn{2}{c||}{2017-0828} & \multicolumn{2}{c}{2017-0129} \\
% & unstructured scene & structured scene & unstructured scene & structured scene & unstructured scene & structured scene \\
 & {2017-0827} & {2017-0828} & {2018-0129} \\
\cline{2-4}
\makecell[r]{Trans Error (m) \\Rota Error (deg)} & \makecell[c]{.25/.5/1.0/5.0 \\ 2/5/8/10} & \makecell[c]{.25/.5/1.0/5.0 \\ 2/5/8/10} & \makecell[c]{.25/.5/1.0/5.0 \\ 2/5/8/10} \\
\hline
1P1L-4DoF & \textcolor{white}{0}{2.1} / \textcolor{white}{0}{5.2} / \textcolor{white}{0}{9.7} / \textcolor{black}{21.5}  &
\textcolor{white}{0}{2.6} / \textcolor{white}{0}{7.6} / \textcolor{black}{14.3} / \textcolor{black}{26.1}  & \textcolor{black}{2.5} / \textcolor{white}{0}{6.0} / \textcolor{black}{10.6} / \textcolor{black}{21.8} \\
\hline
1P1L-6DoF & \textcolor{white}{0}{6.3} / \textcolor{black}{10.2} / \textcolor{black}{15.3} / \textcolor{black}{24.3} & \textcolor{white}{0}{7.7} / \textcolor{black}{14.4} / \textcolor{black}{20.4} / \textcolor{black}{28.6} & \textcolor{blue}{6.0} / \textcolor{white}{0}{9.3} / \textcolor{black}{12.5} / \textcolor{black}{23.7} \\
\hline
2P-4DoF & \textcolor{white}{0}{3.6} / \textcolor{black}{10.0} / \textcolor{black}{17.2} / \textcolor{black}{31.1} & \textcolor{white}{0}{8.6} / \textcolor{black}{19.3} / \textcolor{black}{28.0} / \textcolor{blue}{39.3} & \textcolor{black}{0.4} / \textcolor{white}{0}{5.9} / \textcolor{black}{12.6} / \textcolor{black}{25.4} \\
\hline
2P-6DoF & \textcolor{white}{0}\textcolor{blue}{9.3} / \textcolor{blue}{17.1} / \textcolor{blue}{24.3} / \textcolor{red}{34.4} & \textcolor{red}{15.7} / \textcolor{red}{25.3} / \textcolor{red}{32.5} / \textcolor{red}{39.8} & {5.9} / \textcolor{blue}{11.5} / \textcolor{blue}{16.6} / \textcolor{black}{27.3} \\
\hline
mixed-4DoF & \textcolor{white}{0}{5.7} / \textcolor{black}{14.4} / \textcolor{black}{22.4} / \textcolor{blue}{34.1} & \textcolor{white}{0}{7.1} / \textcolor{black}{17.3} / \textcolor{black}{25.6} / \textcolor{black}{35.5} & \textcolor{black}{3.9} / \textcolor{white}{0}{9.3} / \textcolor{black}{15.6} / \textcolor{red}{28.5} \\
\hline
mixed-6DoF & \textcolor{red}{12.1} / \textcolor{red}{18.7} / \textcolor{red}{24.4} / \textcolor{black}{33.3} & \textcolor{blue}{13.7} / \textcolor{blue}{21.9} / \textcolor{blue}{28.5} / \textcolor{black}{35.1} & \textcolor{red}{7.3} / \textcolor{red}{12.6} / \textcolor{red}{17.0} / \textcolor{blue}{27.7} \\
\hline
P3P & \textcolor{white}{0}{6.7} / \textcolor{black}{14.7} / \textcolor{black}{21.3} / \textcolor{black}{32.2} & \textcolor{black}{10.1} / \textcolor{black}{19.8} / \textcolor{black}{28.2} / \textcolor{black}{37.1} & \textcolor{black}{4.3} / \textcolor{white}{0}{9.7} / \textcolor{black}{12.9} / \textcolor{black}{22.8} \\
\hline
EPnP & \textcolor{white}{0}{7.6} / \textcolor{black}{16.6} / \textcolor{black}{23.5} / \textcolor{black}{32.3} & \textcolor{white}{0}{9.6} / \textcolor{black}{20.3} / \textcolor{black}{28.4} / \textcolor{black}{38.4} & \textcolor{black}{5.3} / \textcolor{black}{11.1} / \textcolor{black}{15.0} / \textcolor{black}{23.8}
\end{tabular}}
\end{center}
\label{table.whole}
\end{table}

\begin{table}[tbp]
\caption{The automatic model selection experiment.}
\begin{center}
\resizebox{0.43\textwidth}{!}{
\begin{tabular}{c||c|c||c|c||c|c||c}
Method & \makecell[c]{1P1L \\ 4DoF} & \makecell[c]{1P1L \\ 6DoF} & \makecell[c]{2P \\ 4DoF} & \makecell[c]{2P \\ 6DoF} & \makecell[c]{mixed \\ 4DoF} & \makecell[c]{mixed \\ 6DoF} & selection \\
\hline
AUC & 0.51 & 0.58 & 0.78 & \textcolor{blue}{0.81} & 0.71 & 0.72 & \textcolor{red}{0.84} \\
\end{tabular}}
\end{center}
\label{table.selection}
\end{table}

%                                  % on the last page of the document manually. It shortens
%                                  % the textheight of the last page by a suitable amount.
%                                  % This command does not take effect until the next page
%                                  % so it should come on the page before the last. Make
%                                  % sure that you do not shorten the textheight too much.

%%%%%%%%%%%%%%%%%%%%%%%%%%%%%%%%%%%%%%%%%%%%%%%%%%%%%%%%%%%%%%%%%%%%%%%%%%%%%%%%

%%%%%%%%%%%%%%%%%%%%%%%%%%%%%%%%%%%%%%%%%%%%%%%%%%%%%%%%%%%%%%%%%%%%%%%%%%%%%%%%

\section*{Acknowledgment}

%This work was supported in part by the National Key R\&D Program of China (2017YFB1300400), in part by the National Nature Science Foundation of China (U1609210).
This work was supported in part by the National Key R\&D Program of China (2017YFC0806501), in part by the Science and Technology Project of Zhejiang Province (2019C01043), and in part by the Science and Technology on Space Intelligent Control Laboratory (HTKJ2019KL502002).

%%%%%%%%%%%%%%%%%%%%%%%%%%%%%%%%%%%%%%%%%%%%%%%%%%%%%%%%%%%%%%%%%%%%%%%%%%%%%%%%

\bibliographystyle{ieeetr} %% setting the cite style
\bibliography{jym}

\addtolength{\textheight}{-12cm}   % This command serves to balance the column lengths

\end{document}